\documentclass{article}
\usepackage{spconf}
\usepackage{amsmath,graphicx}
\usepackage{enumitem}
\setlist{nosep, leftmargin=14pt}

\usepackage{mwe} 
\usepackage{graphicx}
\usepackage{amsmath}
\usepackage{amsfonts}
\usepackage[table]{xcolor}
\usepackage{multirow}
\usepackage{booktabs}
\usepackage{bbm}
\usepackage{url}
\usepackage{hyperref}
\usepackage{float}
\usepackage{adjustbox}
\usepackage{arydshln}
\usepackage[table]{xcolor}
\usepackage[table]{xcolor}
\usepackage[table]{xcolor}
\definecolor{mcolor}{RGB}{25,65,140}
\definecolor{lightmcolor}{RGB}{230,238,250}
\definecolor{mcolor}{RGB}{25,65,140}
\definecolor{lightmcolor}{RGB}{230,238,250}

\definecolor{skyblue}{RGB}{135, 206, 235}

\title{\textit{\underline{AURORA}}: \underline{A}daptive \underline{U}nified \underline{R}epresentation f\underline{O}r \underline{R}obust \underline{U}ltrasound \underline{A}nalysis}

\name{%
\begin{tabular}{@{}c@{}}
Ufaq Khan$^{1}$, L D M S Sai Teja$^{2}$, Ayuba Shakiru$^{3}$, Mai A. Shaaban$^{1}$\\
\textit{Yutong Xie}$^{1}$, Muhammad Bilal$^{3}$, Muhammad Haris Khan$^{1}$
\end{tabular}%
}

\address{$^1$MBZUAI, Abu Dhabi, UAE, $^2$ NIT Silchar, India, 
$^3$Birmingham City University, UK.}

\begin{document}
\maketitle
\begin{abstract}
    Ultrasound images can vary widely across scanners, operators, and anatomical targets, so models trained in one setting often generalize poorly to new hospitals and clinical conditions. The Foundation Model Challenge for Ultrasound Image Analysis (FMC-UIA) reflects this scenario by requiring a single model to handle multiple tasks like segmentation, detection, classification, and landmark regression across diverse organs and datasets. We propose a unified multi-task framework based on a transformer visual encoder from the Qwen3-VL family. Intermediate token features are projected into spatial feature maps and fused using a lightweight multi-scale feature pyramid, enabling both pixel-level predictions and global reasoning within a shared representation. Each task is addressed with a small task-specific prediction head, while training uses task-aware sampling and selective loss balancing to manage heterogeneous supervision and reduce task imbalance. Our method is designed to be straightforward to optimize and adaptable across a wide range of ultrasound analysis tasks. The score improved from \textbf{67\%} to \textbf{85\%} on validation and reached \textbf{81.84\%} average on the official test set across all tasks. The code is publicly available at \href{https://github.com/saitejalekkala33/FMCUIA-ISBI.git}{Github}.

\end{abstract}
\vspace{-5mm}
\section{Introduction}
\label{sec:intro}

Ultrasound is a first-line imaging modality due to its portability, real-time, and free ionizing radiation. At the same time, automated analysis is difficult: appearance changes with probe pose and operator technique; speckle statistics vary across vendors and settings; and clinically relevant targets are usually small, occluded, or weakly contrasted \cite{noble2006ultrasound}. Such that models trained in one setting often degrade when deployed on new devices, institutions, or patient cohorts\cite{khan2025enhancing}. Recent work has moved from task-specific ultrasound models toward generalist representations inspired by foundation models in vision. Promptable segmentation frameworks such as "Segment Anything" shows how large-scale pretraining can transfer across diverse settings~\cite{kirillov2023segment,ravi2024sam}, and "MedSAM" demonstrates the value of domain-tailored pretraining in medical imaging~\cite{ma2024segment}. For ultrasound, modality-specific efforts including UltraSam~\cite{meyer2025ultrasam}, UltraFedFM~\cite{jiang2025pretraining}, and large-scale self-supervised pretraining studies~\cite{ambsdorf2025general} further highlight that scale and diversity can improve robustness. However, most ultrasound foundation models remain single-task, limiting support for masks, labels, boxes, and landmarks in one model.
% However, most existing ultrasound foundation-model works still concentrate on a single task family, leaving open how to support dense masks, image-level labels, bounding boxes, and continuous landmarks within one shared model.

The \emph{Foundation Model Challenge for Ultrasound Image Analysis (FMC-UIA 2026)}
% \footnote{\url{https://www.codabench.org/competitions/11539/}}
~\cite{codabench_fmc_uia} addresses this gap by requiring a single submission to perform segmentation, classification, detection\cite{khan2024ultraweak}, and landmark regression across many organs, with evaluation on unseen domains to stress-test generalization\cite{khan2026calibration}. In this paper, we present a unified multi-task baseline aligned with the challenge protocol and built on a pretrained transformer vision encoder. We use the vision encoder from Qwen3-VL \cite{bai2025qwen3} as backbone and bridge intermediate token representations to multi-scale spatial feature maps through a lightweight feature pyramid, enabling one shared representation to support both dense and global/coordinate objectives. \textbf{Contributions.} We propose (i) a unified multi-task ultrasound framework based on a pretrained \textit{Qwen3-VL} vision encoder, enabling a single model to jointly address segmentation, detection, classification, and landmark regression through a shared representation; (ii) a token-to-map bridging mechanism with multi-scale (FPN-style) feature fusion that converts intermediate transformer token activations into spatial feature maps, supporting both dense prediction and global/coordinate reasoning; and (iii) an efficient task-routing and optimization strategy comprising lightweight task-specific heads with explicit routing, together with task-aware sampling and selective loss balancing to accommodate heterogeneous supervision, alleviate task imbalance, and improve memory and training efficiency.

% The key contributions are as follows:
% \begin{itemize}
%   \item \textbf{Unified multi-task ultrasound framework:} We propose a single model built on a pretrained \textit{Qwen3-VL} vision encoder that jointly addresses segmentation, detection, classification, and landmark regression within a shared representation.

%   \item \textbf{Token-to-map bridge with multi-scale fusion:} We convert intermediate transformer token activations into spatial feature maps and fuse them through a multi-scale pyramid (FPN-style), enabling both dense prediction and global/coordinate reasoning.

%   \item \textbf{Efficient task routing and robust optimization:} We design lightweight task-specific heads with explicit routing, and employ task-aware sampling with selective loss balancing to handle heterogeneous supervision, mitigate task imbalance, and improve memory efficiency.
% \end{itemize}
\begin{figure*}[hbtp]
  \centering
  \includegraphics[width=0.9\textwidth]{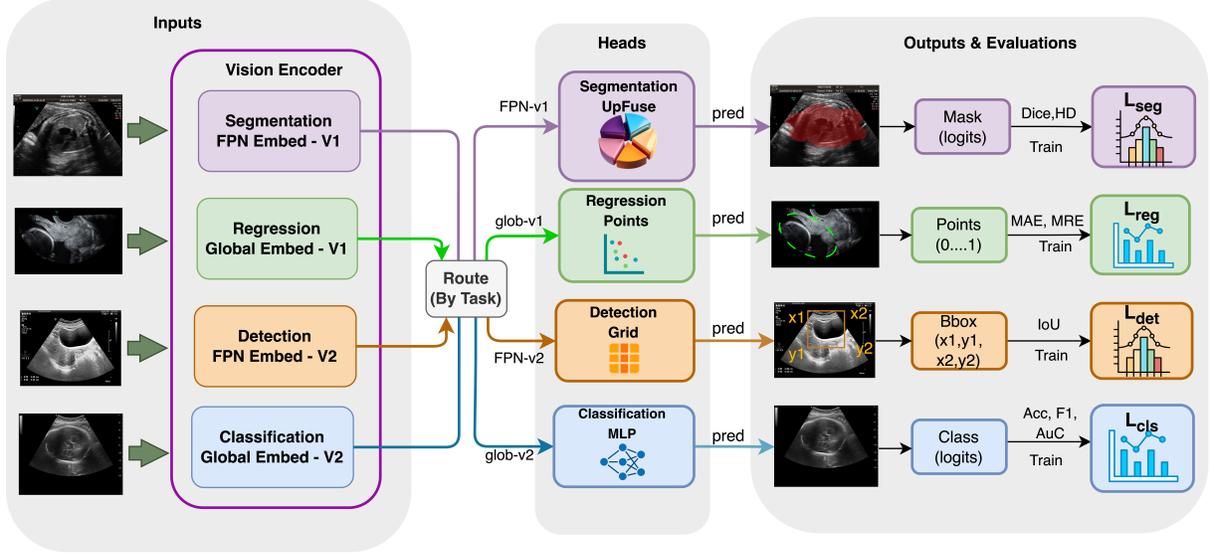}
  \caption{Overview of the proposed unified multi-task framework. Shared Qwen3-VL encoder extracts multi-depth tokens, reshaped and fused into a four-level pyramid. Dense heads use pyramid features; global heads use pooled multi-scale embeddings.}
  \label{fig:chal}
\end{figure*}
\section{Method}
\label{sec:method}
\vspace{-5mm}
\subsection{Overview}
Given an ultrasound image $X\in\mathbb{R}^{H\times W\times 3}$ and task identifier $t$, our model learns a task-conditional predictor $f_{\theta}(X,t)\mapsto \hat{y}_t$, where $\hat{y}_t$ corresponds to segmentation, detection, classification, or landmark regression. The architecture follows a shared private design: a shared visual encoder produces a common representation, which is transformed into a multi-scale feature pyramid, and lightweight task-specific heads are attached. Routing is explicit via $t$, with dense tasks (segmentation/detection) operating on pyramid features and global tasks (classification/regression) using a pooled embedding derived from the pyramid. An overview is shown in Fig.~\ref{fig:chal}.

% \section{Method}
% \label{sec:method}
% \vspace{-5mm}
% \subsection{Overview}
% Let $X \in \mathbb{R}^{H \times W \times 3}$ denote an input ultrasound image, where $H$ and $W$ are the image height and width, respectively, and let $t$ denote the task identifier associated with the sample.
% % be its task identifier provided by the CSV index. 
% The proposed model implements a task-conditional mapping $f_{\theta}:\ (X,t)\ \mapsto\ \hat{y}_t,$
% % \begin{equation}
% % f_{\theta}:\ (X,t)\ \mapsto\ \hat{y}_t,
% % \end{equation}
% where the output type $\hat{y}_t$ is determined by the task family (segmentation, detection, classification, or landmark regression). The architecture follows a shared--private design: a \emph{shared} visual encoder produces a common representation, which is converted into a \emph{multi-scale} feature hierarchy, and four \emph{task-specific} prediction heads are attached. Task routing is explicit, i.e., the identifier $t$ selects the corresponding head. Dense prediction tasks (segmentation and detection) operate on the multi-scale pyramid, whereas global tasks (classification and regression) operate on a compact pooled representation derived from the pyramid. An overview of the unified architecture is shown in Fig.~\ref{fig:chal}.

\subsection{Visual encoder and intermediate representations}
We use the Qwen3-VL vision encoder as backbone and discard the language model to reduce memory. Inputs are resized to a fixed square resolution and preprocessed to normalized pixels with a grid specification; for still images the temporal grid is 1, yielding a spatial lattice of size $h{\times}w$ (grayscale is replicated to 3 channels). To obtain multi-level features, we attach forward hooks to four transformer blocks roughly uniformly spaced in depth, producing token activations $\mathbf{T}^{(\ell)}\!\in\!\mathbb{R}^{B\times N\times C}$. Using $N{=}hw$, we reshape them into spatial maps $\mathbf{F}^{(\ell)}\!\in\!\mathbb{R}^{B\times C\times h\times w}$, ranging from low-level detail to high-level semantic representations.

\subsection{Token pyramid and feature pyramid fusion}
The channel dimensionality of each hooked map $\mathbf{F}^{(\ell)}$ depends on the backbone; therefore, we project all maps to a shared width $D$ using a learnable $1{\times}1$ convolution instantiated lazily on first use. We construct a four-level hierarchy by resizing features to relative scales $\{1,\tfrac{1}{2},\tfrac{1}{4},\tfrac{1}{8}\}$ of the finest level and fuse them with a top-down FPN with lateral connections, yielding pyramid features $\{\mathbf{P}_2,\mathbf{P}_3,\mathbf{P}_4,\mathbf{P}_5\}$ at channel width $D$. Dense heads operate on the multi-scale pyramid, whereas global tasks use a pooled representation obtained by concatenating global average pooled vectors from each level, i.e., $\mathbf{e}=[\mathrm{GAP}(\mathbf{P}_2);\mathrm{GAP}(\mathbf{P}_3);\mathrm{GAP}(\mathbf{P}_4);\mathrm{GAP}(\mathbf{P}_5)]\in\mathbb{R}^{4D}$, which captures both fine-grained detail and coarse contextual information.

% \subsection{Token pyramid and feature pyramid fusion}
% The channel size of each hooked feature map $\mathbf{F}^{(\ell)}$ varies with backbone, so we project all maps to a common width $D$ with a learnable $1\times 1$ convolution instantiated lazily on first use. We then build a four-level hierarchy by resizing the maps to relative scales $\{1,\tfrac{1}{2},\tfrac{1}{4},\tfrac{1}{8}\}$ of the finest leve and fuse them with a top-down FPN with lateral connections, producing pyramid features $\{\mathbf{P}_2,\mathbf{P}_3,\mathbf{P}_4,\mathbf{P}_5\}$ at a common channel width $D$. Dense heads consume these multi-scale features, while global tasks use a pooled embedding formed by concatenating global average pooled vectors from each level \textit{e}, which captures both fine detail and coarse context.
% \begin{multline}
% \mathbf{e} =
% \big[\,\mathrm{GAP}({P}_2)\ ;\ \mathrm{GAP}({P}_3)\ ; \\
% \mathrm{GAP}({P}_4)\ ;\ \mathrm{GAP}({P}_5)\,\big]
% \in \mathbb{R}^{4D}
% \end{multline}
\subsection{Task-specific heads and loss functions}
\noindent\textbf{Segmentation head.}
Segmentation exploits the full multi-scale pyramid. Each level is refined with lightweight residual blocks and channel recalibration, and we apply an Atrous Spatial Pyramid Pooling (ASPP)  module at the coarsest level to enlarge the effective receptive field. A top-down decoder then upsamples and merges features using gated skip connections. A final $1{\times}1$ convolution predicts logits for the maximum number of classes across all segmentation tasks; for a given task, we slice the corresponding channels and upsample predictions to the input resolution. We train the head with a standard weighted cross-entropy combined with a soft Dice term (computed over foreground classes), using $\mathcal{L}_{\mathrm{seg}}=\mathcal{L}_{\mathrm{ce}}+\lambda_{\mathrm{dice}}\mathcal{L}_{\mathrm{dice}}$.

\noindent\textbf{Detection head.}
Detection is implemented as an anchor-free grid predictor operating on the finest pyramid level $\mathbf{P}_2$. To inject additional context, we resize $\mathbf{P}_3$ and $\mathbf{P}_4$ to the $\mathbf{P}_2$ resolution, concatenate them with $\mathbf{P}_2$ and normalized coordinate channels, and feed the resulting tensor to a lightweight convolutional tower. The head predicts, at each grid location, a bounding box $(x_1,y_1,x_2,y_2)$ and an objectness score. Following the challenge protocol, inference selects the box at the location with maximum objectness, avoiding other post-processing. For training, we treat locations whose grid center lies inside the ground-truth box as positives and optimize a focal-weighted binary cross-entropy for objectness together with box regression and an IoU-based penalty, i.e., $\mathcal{L}_{\mathrm{det}}=\mathcal{L}_{\mathrm{obj}}+\lambda_{\mathrm{box}}\mathcal{L}_{\mathrm{box}}+\lambda_{\mathrm{iou}}\mathcal{L}_{\mathrm{iou}}$, where $\mathcal{L}_{\mathrm{box}}$ is SmoothL1 and $\mathcal{L}_{\mathrm{iou}}$ is $(1-\mathrm{IoU})$, both computed over positive locations only.

% \noindent \textbf{Classification head.}
% Classification operates on the pooled embedding $\mathbf{e}$. A projection layer maps $\mathbf{e}$ to a fixed hidden dimension, followed by layer normalization and dropout. We then apply gated MLP residual blocks and a final linear classifier that outputs logits up to the maximum class count, with task-specific slicing as needed. With $C$ classes, the head outputs logits $Z \in \mathbb{R}^{B \times C}$ and $P=\mathrm{softmax}(Z)$. With labels $Y$, the cross-entropy loss with $\alpha$ are optional class weights is 
% \begin{equation}
% \mathcal{L}_{\mathrm{cls}} =
% -\frac{1}{B}\sum_{b=1}^{B} \alpha_{Y_b}\log(P_{b,Y_b}),
% \end{equation}
\noindent\textbf{Classification head.}
Classification is performed on the pooled representation $\mathbf{e}$. We first project $\mathbf{e}$ to a fixed hidden dimension, followed by layer normalization and dropout, then apply a small stack of gated MLP residual blocks. A final linear layer predicts logits for the maximum class count across classification tasks; for each task, we slice the relevant logits. Training uses standard (optionally class-weighted) cross-entropy, i.e., $\mathcal{L}_{\mathrm{cls}}=-\frac{1}{B}\sum_{b=1}^{B}\alpha_{Y_b}\log p_{b,Y_b}$, where $p_{b,Y_b}$ denotes the predicted probability of the ground-truth class and $\alpha$ are optional class weights.

% \noindent \textbf{Regression head.}
% Landmark regression also uses the pooled embedding $\mathbf{e}$. The head outputs $2M_{\max}$ values, and for a task with $M$ landmarks, we take the first $2M$ and apply a sigmoid, so the prediction lie in $[0, 1].$ Ground-truth landmarks are normalized by image $W$, $H$ during loading, so training runs in normalized coordinates; pixel-space errors are computed only for reporting. This predict $M$ landmarks as normalized coordinates in $[0,1]$, stacked as $\hat{r}\in \mathbb{R}^{B \times 2M}$ with targets $r \in \mathbb{R}^{B \times 2M}$. For an error $d=\hat{r}-r$, SmoothL1 with parameter $\beta$ is $\ell_{\mathrm{smooth}}(d).$ We use a combination of SmoothL1 and L1 losses, $\mathcal{L}_{\mathrm{reg}} = \mathcal{L}_{\mathrm{smooth}} + \lambda_{\mathrm{l1}}\mathcal{L}_{\mathrm{l1}}$.

% \begin{multline}
% \mathcal{L}_{\mathrm{smooth}}
% = \frac{1}{2MB}
% \sum_{b=1}^{B}\sum_{k=1}^{2M}
% \ell_{\mathrm{smooth}}(\hat{r}_{b,k}-r_{b,k}), \\
% \mathcal{L}_{\mathrm{l1}}
% = \frac{1}{2MB}
% \sum_{b=1}^{B}\sum_{k=1}^{2M}
% \left|\hat{r}_{b,k}-r_{b,k}\right|.
% \end{multline}

\noindent\textbf{Regression head.}
Landmark regression also operates on the pooled embedding $\mathbf{e}$. The head predicts $2M_{\max}$ values; for a task with $M$ landmarks, we keep the first $2M$ and apply a sigmoid to obtain normalized coordinates in $[0,1]$. Ground-truth landmarks are normalized by image width/height at loading time, so training is performed entirely in normalized coordinates (pixel-space errors are reported only for evaluation). We supervise the regressor using a robust combination of SmoothL1 and $\ell_1$ losses, i.e., $\mathcal{L}_{\mathrm{reg}}=\mathcal{L}_{\mathrm{smooth}}+\lambda_{\mathrm{l1}}\mathcal{L}_{\mathrm{l1}}$, computed as the mean error over all landmark coordinates in the batch.
% \subsection{Optimization and Sampling Strategy}
% To address severe task imbalance, we adopt a temperature-based task sampler that combines per-task frequency with an EMA of recent losses, together with selective loss balancing (Table~\ref{tab:hyperparams}). We resize inputs to $448{\times}448$ (grayscale replicated to 3 channels) and train \texttt{Qwen/Qwen3-VL-8B-Instruct} with an FPN of width $D{=}256$ using bf16 mixed precision, gradient checkpointing, accumulation (effective batch size 16), and gradient clipping. Optimization uses AdamW with weight decay $0.01$ and separate learning rates for heads/backbone. We apply light photometric and blur augmentations, and perform inference with a single forward pass (no TTA; seed 42). Full training, augmentation, and inference settings are reported in Table~\ref{tab:hyperparams}, with implementation details in Table~\ref{tab:impl_env}.
% \newcommand{\qwen}{\texttt{Qwen3-VL-8B}}
% \subsection{Optimization and Sampling Strategy}
% To mitigate task imbalance, we use a temperature-based sampler combining task frequency with a loss-EMA, together with selective loss balancing. We train $\qwen$ (+FPN, $D{=}256$) on $448{\times}448$ inputs (gray$\rightarrow$3ch) with bf16, gradient checkpointing, accumulation (eff.\ batch 16), and gradient clipping. Optimization uses AdamW (wd $0.01$) with separate backbone/head learning rates; we apply light photometric/blur augmentations and perform inference in a single forward pass (no TTA; seed 42). Full settings are in Table~\ref{tab:hyperparams}.
\newcommand{\qwen}{\texttt{Qwen3-VL-8B}}

\subsection{Optimization and Sampling Strategy}
To counter task imbalance, we use a temperature-based sampler mixing task frequency with a loss-EMA (temp.\ 0.7, $\beta{=}0.98$) and dynamic loss reweighting ($\gamma{=}0.5$, clamp $[0.25,4.0]$). We train $\qwen$+FPN ($D{=}256$) on $448{\times}448$ inputs (gray$\rightarrow$3ch) with bf16, gradient checkpointing, accumulation 16 (batch 1; eff.\ 16), and clipping (1.0). AdamW (wd 0.01) uses separate head/backbone LRs ($2{\times}10^{-4}$/$2{\times}10^{-5}$). Augmentations are light (B/C $p{=}0.35$, noise var 5--45 $p{=}0.30$, blur k=3--5 $p{=}0.20$, motion k=3--5 $p{=}0.10$); inference is single-pass (no TTA; seed 42). Experiments run on Ubuntu 24.04 with an NVIDIA L40S (48\,GiB) and AMD EPYC 7R13.

\section{EXPERIMENTS AND RESULTS}
\label{sec:exp}
\subsection{Datasets \& Evaluation Metrics}
FMC-UIA 2026 \cite{codabench_fmc_uia}is a multi-center ultrasound benchmark covering diverse scanners, views, and anatomies, organized into 27 subtasks spanning segmentation, classification, detection, and landmark regression~\cite{lu2022jnu,chen2025comt,zhang2025automatic,chen2024psfhs}. Training and validation was done using the official training split and official validation split drawn from unseen domains to assess cross-domain generalization. The validation set includes 2,674 segmentation samples (12 subtasks), 2,727 classification samples (9 subtasks), 725 detection samples (3 subtasks), and 617 regression samples (3 subtasks). Following the official protocol, we report category-level averages over subtasks: segmentation is evaluated by Dice (DSC) and Hausdorff distance (HD), classification by AUC/F1/MCC, detection by IoU, and regression by mean radial error (MRE) in pixels at the original resolution. We report the official validation results averaged across subtasks within each category: (i) segmentation: mean DSC and mean HD, (ii) classification: mean AUC, mean F1, and mean MCC, (iii) detection: mean IoU, and (iv) regression: mean MRE in pixels.
% Category-averaged final test results are reported in Table~\ref{tab:all_tasks_results_raw} and also summarize ablations across Qwen2.5-VL-7B and Qwen3-VL-8B in Fig.~\ref{fig:ablation}.

\begin{table*}[t!]
\centering
\caption{Unified results table (raw metrics only). Dashes indicate metrics not applicable to that task.}
\label{tab:all_tasks_results_raw}

% extra padding so edge headers aren't clipped
\setlength{\tabcolsep}{6pt}

\resizebox{\linewidth}{!}{%
\begin{tabular}{>{\raggedright\arraybackslash}p{1.7cm}
                >{\raggedright\arraybackslash}p{2.7cm}
                c c cc ccc
                >{\centering\arraybackslash}p{1.7cm}}
\toprule
\rowcolor{mcolor}
\textcolor{white}{\textbf{Task}} &
\textcolor{white}{\textbf{Category}} &
\textcolor{white}{\textbf{IoU}} &
\textcolor{white}{\textbf{MRE}} &
\multicolumn{2}{c}{\textcolor{white}{\textbf{Segmentation}}} &
\multicolumn{3}{c}{\textcolor{white}{\textbf{Classification}}} &
\textcolor{white}{\textbf{Task Score}} \\
& &
\textbf{IoU}$_\text{Raw}$ &
\textbf{MRE}$_\text{Raw}$ &
\textbf{DSC}$_\text{Raw}$ &
\textbf{HD}$_\text{Raw}$ &
\textbf{AUC}$_\text{Raw}$ &
\textbf{F1}$_\text{Raw}$ &
\textbf{MCC}$_\text{Raw}$ & \\
\cmidrule(lr){3-3}\cmidrule(lr){4-4}\cmidrule(lr){5-6}\cmidrule(lr){7-9}
\midrule

Detection & thyroid\_nodule\_det
& 0.717 & -- & -- & -- & -- & -- & -- & 0.920 \\
\midrule

\multirow{3}{*}{Regression} & IUGC
& -- & 47.915 & -- & -- & -- & -- & -- & 0.685 \\
& FUGC
& -- & 17.911 & -- & -- & -- & -- & -- & 0.858 \\
& fetal\_femur
& -- & 71.954 & -- & -- & -- & -- & -- & 0.960 \\
\midrule

\multirow{6}{*}{Segmentation} & cardiac\_multi
& -- & -- & 0.525 & 177.195 & -- & -- & -- & 0.385 \\
& cervix\_multi
& -- & -- & 0.721 & 139.726 & -- & -- & -- & 0.605 \\
& head\_symphysis\_multi
& -- & -- & 0.810 & 43.164 & -- & -- & -- & 0.840 \\
& lung
& -- & -- & 0.857 & 27.683 & -- & -- & -- & 0.866 \\
& cervix
& -- & -- & 0.841 & 16.714 & -- & -- & -- & 0.798 \\
& fetal\_head
& -- & -- & 0.961 & 17.807 & -- & -- & -- & 0.780 \\
\midrule

\multirow{2}{*}{Classification} & fetal\_head\_pos\_cls
& -- & -- & -- & -- & 0.972 & 0.838 & 0.815 & 0.894 \\
& fetal\_sacral\_pos\_cls
& -- & -- & -- & -- & 0.897 & 0.691 & 0.680 & 0.720 \\
\bottomrule
\end{tabular}
}
\end{table*}

\begin{figure}
    \centering
    \includegraphics[width=1.0\linewidth]{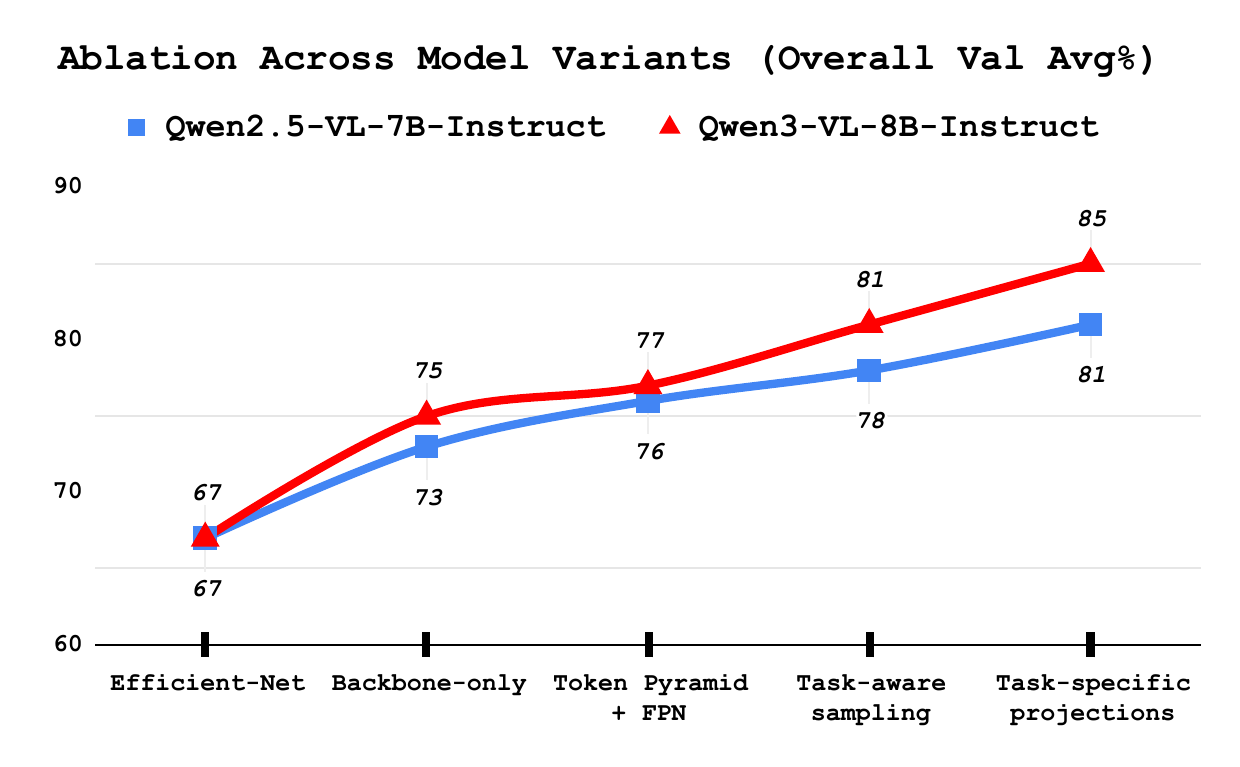}
    \caption{Ablation across model variants, reported the overall average score for the validation split with backbones Qwen2.5-VL-7B-Instruct and Qwen3-VL-8B-Instruct.}
    \label{fig:ablation}
\end{figure}

Table~\ref{tab:all_tasks_results_raw} shows that our model achieves high discrimination in classification subtasks, and thyroid nodule detection attains solid localization performance, indicating that pooled multi-scale embeddings and high-resolution pyramid features are effective for coarse localization. Segmentation performance is heterogeneous: lung and cervix reach strong Dice scores, whereas cardiac multi-class segmentation degrades, reflecting anatomical complexity. Regression exhibits wide variability in mean radial error, suggesting limited geometric fidelity when predicting normalized landmark coordinates from pooled features.

Fig.~\ref{fig:ablation} demonstrates consistent performance gains as architectural components are incrementally introduced. Transforming intermediate transformer tokens into a multi-scale pyramid with FPN fusion yields substantial improvements for dense prediction. Across all configurations, Qwen3-VL-8B consistently outperforms Qwen2.5-VL-7B, confirming that backbone capacity and pretraining quality are primary determinants of cross-task generalization.

\section{Conclusion}
We introduced a unified multi-task model for FMC-UIA that leverages the Qwen3-VL vision encoder and a lightweight token-to-pyramid bridge to transform intermediate transformer tokens into multi-scale spatial feature maps. This enables a single shared representation to support segmentation, detection, classification, and landmark regression through efficient task-specific heads. Overall, the framework is straightforward to optimize and provides competitive performance across heterogeneous objectives; however, it also has limitations, including potential negative transfer between tasks when using one shared backbone, reduced sensitivity in challenging detection scenarios such as very small or ambiguous targets, and possible loss of geometric fidelity due to resizing all inputs to a fixed resolution. Future work will focus on stronger task decoupling (e.g., adapters or routing), scale-aware detection mechanisms, and resolution-aware preprocessing to better preserve fine details and further improve robustness and generalization.
\bibliographystyle{IEEEbib}
\bibliography{ref}

% \clearpage
\appendix
\section{Appendix}

\subsection{Training and Inference Hyperparameters}
We provide a detailed summary of the practical settings used to train and evaluate the unified multi-task model in Table~\ref{tab:hyperparams_appendix}. The table is organized by stage, including task sampling and loss balancing, input preprocessing, optimization, augmentation, and inference. These settings are included to make the experimental pipeline reproducible and to clarify the choices that support stable multi-task learning across heterogeneous objectives. In particular, the reported configuration reflects the balance between memory-efficient large-backbone training (e.g., mixed precision and gradient accumulation) and robust task-level performance (e.g., temperature-based sampling and selective loss weighting).

\begin{table}[H]
\centering
\caption{Training and inference hyperparameters.}
\label{tab:hyperparams_appendix}
\resizebox{\linewidth}{!}{%
\begin{tabular}{lccccc}
\toprule
\rowcolor{mcolor}
\textcolor{white}{\textbf{Section}} &
\multicolumn{2}{c}{\textcolor{white}{\textbf{Hyperparameter}}} &
\multicolumn{3}{c}{\textcolor{white}{\textbf{Value}}} \\
\midrule

\multirow{5}{*}{Task sampling / loss}
& \multicolumn{2}{c}{Temperature} & \multicolumn{3}{c}{$0.7$} \\
& \multicolumn{2}{c}{Loss EMA $\beta$} & \multicolumn{3}{c}{$0.98$} \\
& \multicolumn{2}{c}{Dynamic weighting $\gamma$} & \multicolumn{3}{c}{$0.5$} \\
& \multicolumn{2}{c}{Weight clamp} & \multicolumn{3}{c}{$[0.25,\ 4.0]$} \\
& \multicolumn{2}{c}{Sampler type} & \multicolumn{3}{c}{\shortstack{Temperature-based \\(frequency + EMA loss)}} \\
\midrule

\multirow{1}{*}{Input / preprocessing}
& \multicolumn{2}{c}{Image resize} & \multicolumn{3}{c}{$448 \times 448$} \\
\midrule

\multirow{6}{*}{Training}
& \multicolumn{2}{c}{Precision} & \multicolumn{3}{c}{bf16 mixed precision} \\
& \multicolumn{2}{c}{Gradient checkpointing} & \multicolumn{3}{c}{Enabled} \\
& \multicolumn{2}{c}{Per-device batch size} & \multicolumn{3}{c}{$1$} \\
& \multicolumn{2}{c}{Gradient accumulation} & \multicolumn{3}{c}{$16$ steps} \\
& \multicolumn{2}{c}{Effective batch size} & \multicolumn{3}{c}{$16$} \\
& \multicolumn{2}{c}{Gradient clipping} & \multicolumn{3}{c}{Max norm $1.0$} \\
\midrule

\multirow{3}{*}{Optimization}
& \multicolumn{2}{c}{Optimizer} & \multicolumn{3}{c}{AdamW} \\
& \multicolumn{2}{c}{Weight decay} & \multicolumn{3}{c}{$0.01$} \\
& \multicolumn{2}{c}{LR (heads, backbone)} & \multicolumn{3}{c}{$2\times 10^{-4}$, $2\times 10^{-5}$} \\
\midrule

\multirow{4}{*}{Augmentations}
& \multicolumn{2}{c}{Brightness/contrast} & \multicolumn{3}{c}{$p=0.35$} \\
& \multicolumn{2}{c}{Gaussian noise} & \multicolumn{3}{c}{$\texttt{var\_limit}=5$--$45,\ p=0.30$} \\
& \multicolumn{2}{c}{Gaussian blur} & \multicolumn{3}{c}{Kernel $3$--$5,\ p=0.20$} \\
& \multicolumn{2}{c}{Motion blur} & \multicolumn{3}{c}{Kernel $3$--$5,\ p=0.10$} \\
\midrule

\multirow{1}{*}{Inference}
& \multicolumn{2}{c}{Random seed} & \multicolumn{3}{c}{$42$} \\
\bottomrule
\end{tabular}
}
\end{table}

\subsection{Implementation and Environment Details}
To support transparency and reproducibility, Table~\ref{tab:impl_env_appendix} documents the implementation and hardware/software environment used in our experiments. We report the operating system, programming language, accelerator and memory configuration, processor details, and the core training framework stack. These details help contextualize the reported runtime behavior and provide a practical reference for researchers who aim to reproduce or extend the presented multi-task ultrasound pipeline under comparable compute settings.

\begin{table}[!htbp]
\centering
\caption{Implementation and environment details.}
\label{tab:impl_env_appendix}
\resizebox{\linewidth}{!}{%
\begin{tabular}{lc}
\toprule
\rowcolor{mcolor}
\textcolor{white}{\textbf{Component}} & \textcolor{white}{\textbf{Setting}} \\
\midrule
Operating System     & Ubuntu 24.04.4 LTS\\
Programming Language & Python \\
GPU                  & NVIDIA L40S \\
VRAM                 & 48GiB \\
% \textbf{Code}        & (repository link) \\
Processor            & AMD EPYC 7R13 Processor \\
Training Framework   & Hugging Face Transformers, Accelerate \\
\bottomrule
\end{tabular}
}
\end{table}

\subsection{Feature Routing and Prediction Heads}
The backbone records four intermediate transformer activations, reshapes them to spatial maps using the processor-provided grid, and exposes two feature pathways. The first pathway constructs a token pyramid followed by an FPN and is used for detection and classification. The second pathway directly pools/downsamples the hooked feature maps and is used for segmentation and regression. This routing and the heads predicts task-specific outputs by slicing from the maximum output dimensionality across tasks.

\begin{table}[H]
\centering
\caption{Task routing and prediction heads.}
\label{tab:appendix_heads}
\footnotesize
\setlength{\tabcolsep}{4pt}
\begin{tabular}{p{1.7cm}p{1.5cm}p{3.9cm}}
\toprule
Task & Features & Head details \\
\midrule
Segmentation & \texttt{fpn\_v1} & \texttt{SegmentationHeadV1}: $P_5$ projection, three bilinear up-fuse blocks, dropout, and a $1{\times}1$ classifier. Logits are sliced to the task-specific class count and upsampled to the input image size. \\
Classification & \texttt{global} & \texttt{ClassificationHead}: dynamic linear projection to 1024 dimensions, layer normalization, GELU, two gated-MLP residual blocks, and a final linear classifier. \\
Regression & \texttt{global\_v1} & \texttt{RegressionHeadV1}: dynamic linear projection to 1024 dimensions, layer normalization, GELU, dropout, and a linear regressor. The first $2M$ outputs are retained and passed through a sigmoid. \\
Detection & \texttt{fpn} & \texttt{DetectionGridHead}: uses $P_2$ together with upsampled $P_3$ and $P_4$, plus normalized coordinate channels. A residual convolutional tower predicts four box channels and one objectness channel per grid cell. \\
\bottomrule
\end{tabular}
\end{table}

\subsection{Supplementary Normalized Results}
Table~\ref{tab:all_tasks_results_norm_appendix} reports the normalized metrics for the same representative task subset shown with raw metrics in Table~\ref{tab:all_tasks_results_raw} of the main paper.

\begin{table}[t]
\centering
\caption{Unified results table (normalized metrics only). Dashes indicate metrics not applicable to that task.}
\label{tab:all_tasks_results_norm_appendix}
\setlength{\tabcolsep}{6pt}
\resizebox{\linewidth}{!}{%
\begin{tabular}{>{\raggedright\arraybackslash}p{1.7cm}
                >{\raggedright\arraybackslash}p{2.7cm}
                c c cc ccc
                >{\centering\arraybackslash}p{1.7cm}}
\toprule
\rowcolor{mcolor}
\textcolor{white}{\textbf{Task}} &
\textcolor{white}{\textbf{Category}} &
\textcolor{white}{\textbf{IoU}} &
\textcolor{white}{\textbf{MRE}} &
\multicolumn{2}{c}{\textcolor{white}{\textbf{Segmentation}}} &
\multicolumn{3}{c}{\textcolor{white}{\textbf{Classification}}} &
\textcolor{white}{\textbf{Task Score}} \\
& &
\textbf{IoU}$_\text{Norm}$ &
\textbf{MRE}$_\text{Norm}$ &
\textbf{DSC}$_\text{Norm}$ &
\textbf{HD}$_\text{Norm}$ &
\textbf{AUC}$_\text{Norm}$ &
\textbf{F1}$_\text{Norm}$ &
\textbf{MCC}$_\text{Norm}$ & \\
\cmidrule(lr){3-3}\cmidrule(lr){4-4}\cmidrule(lr){5-6}\cmidrule(lr){7-9}
\midrule
Detection & thyroid\_nodule\_det
& 0.920 & -- & -- & -- & -- & -- & -- & 0.920 \\
\midrule
\multirow{3}{*}{Regression} & IUGC
& -- & 0.685 & -- & -- & -- & -- & -- & 0.685 \\
& FUGC
& -- & 0.858 & -- & -- & -- & -- & -- & 0.858 \\
& fetal\_femur
& -- & 0.960 & -- & -- & -- & -- & -- & 0.960 \\
\midrule
\multirow{6}{*}{Segmentation} & cardiac\_multi
& -- & -- & 0.295 & 0.474 & -- & -- & -- & 0.385 \\
& cervix\_multi
& -- & -- & 0.426 & 0.784 & -- & -- & -- & 0.605 \\
& head\_symphysis\_multi
& -- & -- & 0.744 & 0.937 & -- & -- & -- & 0.840 \\
& lung
& -- & -- & 0.806 & 0.926 & -- & -- & -- & 0.866 \\
& cervix
& -- & -- & 0.746 & 0.849 & -- & -- & -- & 0.798 \\
& fetal\_head
& -- & -- & 0.753 & 0.806 & -- & -- & -- & 0.780 \\
\midrule
\multirow{2}{*}{Classification} & fetal\_head\_pos\_cls
& -- & -- & -- & -- & 0.866 & 0.904 & 0.914 & 0.894 \\
& fetal\_sacral\_pos\_cls
& -- & -- & -- & -- & 0.554 & 0.804 & 0.803 & 0.720 \\
\bottomrule
\end{tabular}
}
\end{table}

\subsection{Optimization, Sampling, and Runtime Settings}
We formed task-homogeneous mini-batches and samples task identities with a temperature-based distribution. For classification and detection only, we applied an additional dynamic weighting followed by uncertainty-based balancing. Segmentation and regression are optimized with their raw task losses.

\subsection{Detailed loss functions}

\noindent\textbf{Segmentation.}
For a segmentation task with $C$ classes, let $Z \in \mathbb{R}^{B \times C \times H \times W}$ be the predicted logits, $P=\mathrm{softmax}(Z)$, and $G$ the one-hot encoding of the ground-truth mask. Segmentation uses $\mathcal{L}_{\mathrm{seg}} = \mathcal{L}_{\mathrm{ce}} + \mathcal{L}_{\mathrm{dice}},$ with

\begin{equation}
\mathcal{L}_{\mathrm{ce}}
=
-\frac{1}{BHW}
\sum_{b,u,v}\sum_{c=0}^{C-1}
w_c\,G_{b,c,u,v}\log P_{b,c,u,v},
\end{equation}
where $w_0=0.2$ for the background class and $w_c=1$ for all foreground classes, and

\begin{multline}
\mathcal{L}_{\mathrm{dice}}
= 1 - \frac{1}{\max(1, C-1)} \\
\sum_{c=1}^{C-1}
\frac{
2 \sum_{b,u,v} P_{b,c,u,v} G_{b,c,u,v} + \varepsilon
}{
\sum_{b,u,v} P_{b,c,u,v}
+ \sum_{b,u,v} G_{b,c,u,v}
+ \varepsilon
}.
\end{multline}

\noindent\textbf{Classification.}
For image-level classification, the released code uses standard cross-entropy without an additional class-weighting term:
\begin{equation}
\mathcal{L}_{\mathrm{cls}}
=
-\frac{1}{B}\sum_{b=1}^{B}\log p_{b,Y_b},
\end{equation}
where $p_{b,Y_b}$ is the predicted probability of the ground-truth class.

\noindent\textbf{Regression.}
For a regression task with $M$ landmarks, the head predicts a vector $\hat{r}\in[0,1]^{B \times 2M}$ after sigmoid activation. Ground-truth coordinates are normalized by the original image width and height during data loading. Regression supervises with SmoothL1 only:
\begin{equation}
\mathcal{L}_{\mathrm{reg}}
=
\frac{1}{B(2M)}
\sum_{b=1}^{B}\sum_{d=1}^{2M}
\ell_{\mathrm{smooth}}(\hat{r}_{b,d}-r_{b,d}),
\end{equation}
where $\ell_{\mathrm{smooth}}$ denotes SmoothL1 with $\beta=1.0$.

\noindent\textbf{Detection.}
The released detector predicts, at each location $(u,v)$ on a feature map of size $H \times W$, a bounding box $\hat{b}_{u,v}=(\hat{x}_1,\hat{y}_1,\hat{x}_2,\hat{y}_2)$ and an objectness logit $\hat{o}_{u,v}$. Before loss computation, the four box channels are passed through a sigmoid and reordered with elementwise min/max so that $\hat{x}_1 \le \hat{x}_2$ and $\hat{y}_1 \le \hat{y}_2$. Grid-center coordinates are defined by
\begin{equation}
g_x(u,v)=\frac{v+0.5}{W},
\qquad
g_y(u,v)=\frac{u+0.5}{H}.
\end{equation}
Given a normalized ground-truth box $b^\star=(x_1^\star,y_1^\star,x_2^\star,y_2^\star)$, the positive set is $\mathcal{P}.$ The objectness loss is weighted BCE with logits is $\mathcal{L}_{\mathrm{obj}}.$
\begin{equation}
\mathcal{L}_{\mathrm{obj}}
=
\operatorname{BCE}_{\mathrm{logits}}(\hat{o},y;\rho),
\qquad
\rho = \operatorname{clip}\!\left(\frac{N_-}{N_+},1,50\right),
\end{equation}
where $N_+$ and $N_-$ denote the number of positive and negative grid cells for the current sample. Box regression and IoU losses are computed only over $\mathcal{P}$, and the total detection loss is $\mathcal{L}_{\mathrm{det}}.$
\begin{equation}
\mathcal{L}_{\mathrm{box}}
=
\frac{1}{|\mathcal{P}|}
\sum_{(u,v)\in\mathcal{P}}
\operatorname{SmoothL1}_{\beta=0.05}(\hat{b}_{u,v},b^\star),
\end{equation}
\begin{equation}
\mathcal{L}_{\mathrm{iou}}
=
\frac{1}{|\mathcal{P}|}
\sum_{(u,v)\in\mathcal{P}}
\left(1-\operatorname{IoU}(\hat{b}_{u,v},b^\star)\right),
\end{equation}

\begin{equation}
\mathcal{L}_{\mathrm{det}}=\mathcal{L}_{\mathrm{obj}}+\lambda_{\mathrm{box}}\mathcal{L}_{\mathrm{box}}+\lambda_{\mathrm{iou}}\mathcal{L}_{\mathrm{iou}}.
\end{equation}

\subsubsection{Uncertainty-based reweighting for selected tasks}
For classification and detection, we additionally apply an uncertainty-style reweighting with a learnable scalar $s_k$ for each task family $k$. We define a positive scale $\sigma_k = 1 + \mathrm{softplus}(s_k),$ and transform the loss as $\tilde{\mathcal{L}}_k.$ This yields bounded, learnable task-family weights that can adapt during training.
\begin{equation}
\tilde{\mathcal{L}}_k = \frac{1}{2\sigma_k^2}\mathcal{L}_k + \log(\sigma_k).
\end{equation}

\end{document}